**TrTr: A Versatile Pre-Trained Large Traffic Model based on Transformer for Capturing Trajectory Diversity in Vehicle Population**

**Ruyi Feng**
Ph. D Candidate
School of Transportation, Southeast University
2 Si Pai Lou, Nanjing, 210096, China
Email: fengruyi@seu.edu.cn

**Zhibin Li, Ph. D., Corresponding Author**
Professor
School of Transportation, Southeast University
2 Si Pai Lou, Nanjing, 210096, China
Tel: +86-13952097374; Fax: +86-25-52091255
Email: lizhibin@seu.edu.cn

**Bowen Liu**
Eng. D Candidate
School of Transportation, Southeast University
2 Si Pai Lou, Nanjing, 210096, China
Email: sivanliu0227@gmail.com

**Yan Ding**
Engineer
Sports and Health Research Institute, Southeast University
2 Si Pai Lou, Nanjing, 210096, China
Email: yandy.ding@gmail.com

Word Count: 6053 words + 2 table (250 words per table) = 6,553 words

*Submitted [9/22/2023]*

*Ruyi Feng et.al.*

**ABSTRACT**

Understanding trajectory diversity is a fundamental aspect of addressing practical traffic tasks. However, capturing the diversity of trajectories presents challenges, particularly with traditional machine learning and recurrent neural networks due to the requirement of large-scale parameters. The emerging Transformer technology, renowned for its parallel computation capabilities enabling the utilization of models with hundreds of millions of parameters, offers a promising solution. In this study, we apply the Transformer architecture to traffic tasks, aiming to learn the diversity of trajectories within vehicle populations. We analyze the Transformer's attention mechanism and its adaptability to the goals of traffic tasks, and subsequently, design specific pre-training tasks. To achieve this, we create a data structure tailored to the attention mechanism and introduce a set of noises that correspond to spatio-temporal demands, which are incorporated into the structured data during the pre-training process. The designed pre-training model demonstrates excellent performance in capturing the spatial distribution of the vehicle population, with no instances of vehicle overlap and an RMSE of 0.6059 when compared to the ground truth values. In the context of time series prediction, approximately 95% of the predicted trajectories' speeds closely align with the true speeds, within a deviation of 7.5144m/s. Furthermore, in the stability test, the model exhibits robustness by continuously predicting a time series ten times longer than the input sequence, delivering smooth trajectories and showcasing diverse driving behaviors. The pre-trained model also provides a good basis for downstream fine-tuning tasks. The number of parameters of our model is over 50 million.

**INTRODUCTION**

Understanding trajectory diversity is essential for addressing practical traffic tasks. Varying trajectories combine the diversity of traffic conditions, resulting in congestion, free flow, synchronized flow, traffic oscillation (*1-3*), as well as other complex phenomena like capacity drop, the emergence, propagation, and dissipation of traffic waves (*4-5*). In them, diverse operational patterns of trajectories, such as car-following, lane-changing, and merging in traffic sections, may lead to disturbances or bottlenecks. If the diversity of trajectories is ignored, it is difficult for researchers to reproduce such traffic conditions in experiments.

The foundation of studying trajectory diversity lies in the establishment of physical models, typical example being modeling car-following behavior. Based on Newell's car-following model (*6*), Chen used the response mode $\tau, \eta$ to increase trajectory diversity and reproduce the generation of motion waves (*7*). Further, the heterogeneity is considered an important property in driving response patterns (*8*). However, the parameters of these physical models are calibrated based on trajectory data and can only reflect the average behavior of specific trajectory groups. The effectiveness of the calibration is affected by the richness of the trajectory containing the driving modes and the ability of the model itself to demonstrate the driving modes (*9*). To represent a more diverse set of trajectories, a corresponding increase in the number of parameters is required. Nevertheless, the labor-intensive calibration work limits the scalability of parameters, making it challenging to encounter physical trajectory models in the traffic field with dozens of parameters.

With the rise of machine learning, algorithms have been applied for trajectory calibration and learning of trajectory characteristics. Applications include trajectory prediction and completion. For instance, Liu et al. (*10*) employed TD Tree to construct the feature tree of traffic flow for trajectory completion. SVM was applied to learning the non-linear lane-changing decision (*11*). In Hu's work, multiply machine learning methods like decision tree, random forest, and gradient boosting decision tree, are applied for real-time evaluation of trajectory rare-end incidents (*12*). However, while machine learning overcomes the difficulty of fitting a large number of parameters, its adaptability to trajectory variations remains insufficient due to the manual selection of training features and the limitations of a single specific task.

Deep learning can accommodate regression modeling with a large number of parameters and can autonomously learn and identify effective features. On time-series data such as traffic, recurrent neural networks play a significant role. In which, techniques like LSTM have been used for trajectory prediction. Mo et. al, (*13*) developed physics-informed deep learning car-following model and applied in LSTM net. Ma and Qu (*14*) modeled car-following behavior into a seq-to-seq LSTM net, applying in platoon simulation. But these time-series prediction methods often fail to fully capture the internal relationships within the variables, which do not align with the spatiotemporal nature of traffic data. Additionally, the recurrent structure of LSTM significantly increases computational load, making it difficult to scale the model for finer-grained scenarios.

Transformer (*15*), centered around the self-attention mechanism, excels at capturing internal dependencies within sequences and, thanks to its massive parameter space, learns fine-grained distinctions. Currently, the most advanced models in the NLP field, such as GPT (*16*), BERT (*17*), BART (*18*), T5 (*19*), are built on the Transformer backbone. Drawing parallels between NLP and traffic-related tasks, the Transformer's success in understanding language structures finds a compelling analogy in the context of trajectories. Vehicle trajectories, akin to sentences' organizational structure, present crucial patterns of interactions between vehicles, resembling the influence of sentence components in language. By leveraging the Transformer's attention mechanism, traffic-related tasks can effectively harness trajectory dependencies and uncover intricate traffic flow patterns. The applicability of Transformer-based models, which has revolutionized NLP, holds the potential to transform traffic analysis and decision-making, enhancing the understanding and management of complex traffic scenarios.

In the case of vast parameter space, accurately training the model to learn trajectory patterns poses a significant challenge. This necessitates careful design of data structures, training methodologies, to fit several downstream tasks (*20*). With the encoder-decoder structure used in Transformers, the model primarily learns the transformation patterns from questions to answers, and the vast parameter combinations could lead to multiple transformation modes for a single objective. To achieve comprehensive generalization and avoid catastrophic forgetting, a well-designed training approach and downstream tasks are necessary to induce the desired transformation patterns. For example, in natural language tasks, the training approaches of models like GPT (*16*), BART (*17*) and T5 (*19*) have proven effective, including adding noise to input sequences during training to deepen the model's understanding of language structures and utilizing masking techniques. Moreover, noise injection techniques have also been extended to image learning, where they have been empirically confirmed to be effective in enhancing model performance (*21-22*).

In this paper, we carefully analyze the adaptability of trajectory data structures and task objectives, draw inspiration from the task processing patterns in the natural language domain, and seamlessly integrate the Transformer's attention mechanism into the working structure to handle sequence-related and inter-sequence correlations in traffic data. We propose a corresponding training mode, enabling Transformers to learn the operational patterns of groups of vehicle trajectories and apply them to various traffic tasks, such as prediction and completion. Section Two introduces the basic structure of the network and task adaptation, Section Three provides details of the experimental design, Section Four presents the model results, and Section Five concludes with a summary and discussion.

## METHODS
### Transformer Architecture

Attention mechanism performs the basic calculation module in Transformer, mainly works by three key vectors: query ($Q$), key ($K$) and value ($V$). Attention weights are computed based on the $Q$ vector, which represents the current state of the model, and a set of ($K, V$) pairs, which represent the input. Their core calculation formation can be seen in **Equation 1**. Where $Q, K, V \in R^{S \times d_k}$ are length-S normalized queries, keys and values of $d_k$-dimension respectively, and Softmax($\cdot$) is conducted row by row. For dealing arrangement and collocation of units in sequence, the self-attention is further developed for extract relationship inside. In self-attention, $Q, K$ and $V$ vector are all equal to input sequence. Besides, a unit may have multiple connections to other components. These connections are searched by several parallel self-attention nets extract independently, calculating various correlation matrixes, and combining them by fusion matrix, called as multi-head attention mechanism. Multi-head attention, innovation of the transformer, has performed an excellent capability in learning inter-cross correlation in natural language processing (NLP) tasks. Specifically, its attention regulates the collocation of adjective with noun to avoid improper combination. Applying in locally trajectory distribution learning, this attention mechanism is guided to generate each vehicle's position with considering vehicle gap and interaction of locality.

$$\text{Attn}(Q, K, V) = Softmax\left(\frac{QK^T}{\sqrt{d_k}}\right)V \tag{1}$$

The above mentioned multi-head attention mechanism forms Transformer net by layers' concatenation, as shown in **Figure 1**. Practically, sequential data go through specialized embedding adapted with tasks, output a feature sequence of specific dimensions. This feature then passes into two main modules, encoder and decoder. Each of them contains a stack of blocks, every block includes multi-head attention nets and a feed-forward section. The decoder distinguishes from encoder by an extra masked multi-head attention nets, ensuring sequential input without answer that should have been predicted. In overall operation, Transformer receives vectors $X$ from embedding module, then send the $X$ vector into encoder and decoder separately. In such processing, encoder can be seen as feature extractor, resolves input vectors $X$ as $Q, K, V$ by layers of multi-head attention and forms a new feature vector. The formed new vector is further sent to decoder as $K, Q$, combining the former embedded vector $X$ as $V$. The decoder nets are responsible for applying the extracted features ($K, Q$) with a prompt sequence ($V$) to complete the final output. During training, encoder for extracting features from $X$ pursue more significant transformation making the out features effective, while decoder need to adapt for correct transformation of the effective features. The parameters in encoder-decoder structure are adjusted automatically in training referring the feedback from *loss*, which indicates difference between outputs and task targets.

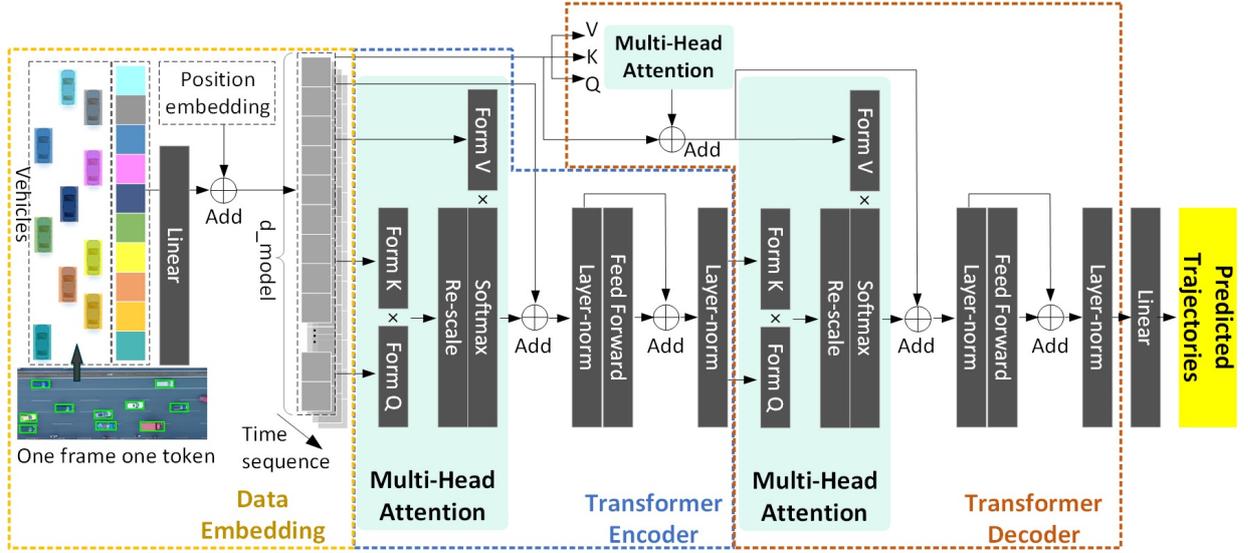

**Figure 1 Framework of Transformer Structure**

**Specialized Embedding**

*Sequential Data Structure Arrangement*

Integrated trajectory data continue for long duration, covering extensive section. Conversely, the motion in split of trajectory, only relates with or be affected by location in a finite spatial-temporal domain, referred to as effective domain. Effective domain contains a group of fragmented trajectories are consecutive or adjacent, and maintain short duration to avoid separation of group members. From this, the integrated trajectories are divided into spatial domains bounded by a maximum length of L and a temporal range of T. Each sub-trajectory within the effective domain is ensured to be continuous and prepared for normalization.

The normalization process involves standardizing the coordinates of the vehicles and the size of their bounding boxes. First, the trajectory is scaled to meters. Then, the coordinate values are divided by the maximum value within the domain, while the vehicle's bounding box is scaled according to the corresponding limit values. For instance, the length of the vehicle does not exceed 20m, and the width does not exceed 4m. Consequently, re-scaled sub-trajectories fall in the interval of 0 to 1.

The preprocessed sub-trajectory conforms to the sequence but follows two dimensions encompassing both temporal and spatial. In order to characterize additional dimensions in sequence data, the task is designed with a specific organized data structure. **Figure 2** illustrates the components of the data, where the vehicle positions, serving as the minimum unit (marked as $V_1, V_2, V_3, ..., V_n$), include box position and box size (X, Y, W, H). This sequence, set parallel with constant but disorderly arrangement in a single frame, aims to represent the spatial distribution. At any given moment, for an example, the boxes with 4 dimension of each, concatenating a sequence in shape of $(1, 4 \times S)$, where the S indicate vehicle counts contained in the section of this moment. We define maxS to limit the scale of vector, not only for easier calculating but also corresponding to the vehicle group members limit. Furthermore, to establish temporal connections, multiple frames equally spaced along time are selected and serialized. Each frame is associated with a frame mark in the condition dimension, representing its relative time within the sequence. The above mentioned structured trajectory data then be dealt to adapt to tasks. Detailed dealing approaches will be expanding on the following sub-sections.

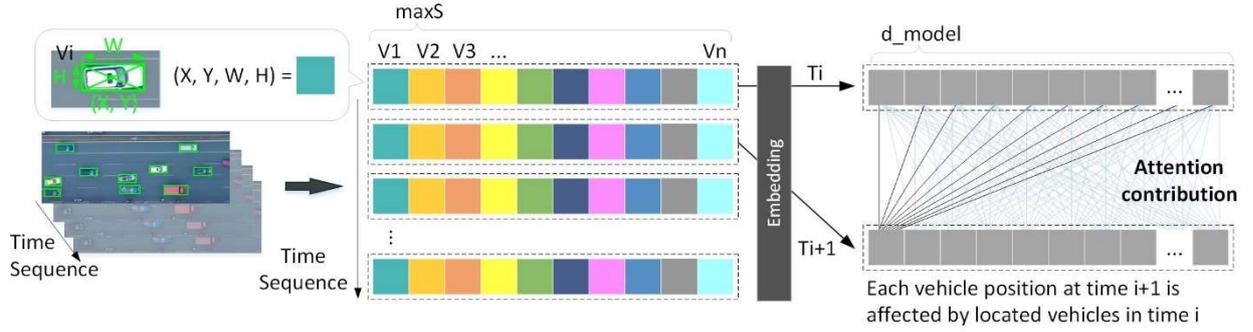

**Figure 2 Components of structured data and attention contribution**

Subsequently, embedding is applied to the structured data as a form of feature pre-extraction. In numerical regression tasks, the linear layer is commonly used for numerical embedding. In the trajectory learning context described above, a numerical embedding is initially applied to the bounding boxes within the trajectories. This process transforms the $(1, 4 \times maxS)$ sequence into an $(1, d_{model})$ sequence as a token using a linear layer or one-dimensional convolution. The position embedding consistent with original Transformer are arranged by corresponding frame mark sequence. The design of the embedding is tailored to effectively cooperate with the data structure to better learn spatiotemporal features. For the attention mechanism work between tokens to search correlation. In this design, temporal sequence is dominant, location distribution of last frame conducts positions of every vehicle in current frame. The distribution positions of all cars in this frame form a spatial association, which is implicitly learned by the attention mechanism. Finally, the tokens and position embedding vectors are combined by addition and then fed into the encoder and decoder components.

*Task Decompose*

By analyzing every task's mechanism, we believe that the prediction task can learn the characteristics of the trajectory more comprehensively. For instance, popular traffic practices such as trajectory compensation, prediction, simulation and traffic control. Trajectory compensation involves bridging the gap between two known points, form a bidirectional prediction task. Trajectory prediction entails supplementing future points based on past information, and it needs to organize the distribution and spatial relationship between vehicles in order to achieve accurate prediction. Traffic simulation initiates with an initial state of a designated area, and the model continuously generates vehicle distribution at subsequent moments, which can be viewed as an extended scenario of trajectory prediction. Similarly, traffic control is also a trajectory prediction task, which adds the influence of a control signal to the original trajectory prediction. Therefore, when choosing a pre-training task, it is most appropriate to use the prediction process

The design of the detail of predict task plays a crucial role in determining the model's scale and the generality of its results. Complex and detailed tasks necessitate more layers and increased model dimensions to capture finer feature details. However, meeting these demands requires scaling up the model parameters, which presents significant challenges in training. In addition to ensuring convergence through sufficiently large batch sizes and avoiding overfitting with rich datasets, the training task itself is closely intertwined with parameter training. It is conceivable that for a specific dataset, when the parameter scale is large enough, numerous combinations of fitting parameters can yield similar results that match the dataset. However, only a few of these combinations satisfy the broader function. To achieve a model with strong transferability, capable of addressing multiple types of tasks, careful formulation of the training tasks is essential to guide parameter regression. Typically, this formulation is carried out through embedding techniques.

Trajectory learning serves as the fundamental objective of our model, seeking to automatically generate the distribution of vehicles in the spatiotemporal domain. To achieve this, vehicles must be placed in realistic spatial locations and exhibit interaction patterns with each other. Furthermore, the model should preserve the natural motion of vehicles in a chronological manner. Given the complexity of designing the entire task, it becomes necessary to decompose it into two independent sub-goals, each addressing a specific aspect. (i)Distribution: This sub-goal focuses on learning the spatial locations of vehicles within a group. By understanding the distribution patterns, the model gains insights into how vehicles are positioned in the spatiotemporal domain. This includes understanding how vehicles influence each other's trajectories and incorporating this interaction information into the prediction process, such as no overlap bounding box. (ii)Temporal: Ensuring the accurate prediction of each time frame while maintaining a consistent chronological sequence is the objective of this sub-goal. It aims to capture the temporal dynamics and dependencies between consecutive time steps. It also responsible for capturing the natural motion characteristics of vehicles to generate realistic trajectories.

*Additional Noise for Embedding*

The sub-goals mentioned earlier are accomplished through additional noise in embedding module. This approach is inspired by the work of Lewis et al. (*18*) in the field of NLP. Their research demonstrated that incorporating noise patterns relevant to specific tasks enhances a model's ability to learn the essential features required for those tasks. Taking this insight into consideration, we have designed tailored noise patterns in conjunction with the characteristics of traffic data to facilitate the realization of the sub-goals. Two corresponding noise are designed to the sub-goals, and are randomly applied in the process of model training to enhance the adaptability of the model to tasks.

- The first type of noise introduced in the embedding module corresponds to the sub-task (i)Distribution. In this type of noise, all vehicles in selected frames are replaced with [MASK] elements. Each selected frames are distributed one [MASK]. The model is then tasked with determining the number of vehicles associated with each [MASK] element and generating their corresponding distributions. To obtain this noise pattern, a sequence mask approach is employed, wherein only a given small portion of the sequence is masked. The position and length of the mask are determined according to a Poisson distribution. When generating a mask, the first step involves obtaining the total length of the mask that satisfies the desired missing rate within the sequence. Subsequently, the probability density function of the Poisson distribution is employed to generate masked length segments that can sum up to the total length. Finally, these masked length segments are randomly distributed throughout the sequence.
- Second, sequence is reorganized by exchanging two frames. A pair of frames participant in switch is randomly chosen, regardless of its adjacent or with a span. Then, vehicles in this pair of frames will be swapped and also its corresponding frame embedding swapped. According this, model is trained to identify chronological sequence and revise errors, realizing the (ii)Temporal sub-task.

**EXPERIMENT DESIGN**

In our experiment, we tailored the decisive parameters of the Transformer training to align with the characteristics of the traffic data. Due to the fact that vehicle behavior is predominantly influenced by the spatial-temporal domain, the sequence length does not become excessively long compared to NLP tasks. Considering a maximum of 10 vehicles in a group, we selected their trajectory points across 20 consecutive frames. Therefore, we set the maximum sequence length to be 20. For the minimum unit, which is the bounding box of the vehicles, comprising 4 dimensions (X, Y, W, H), we expanded it to the dimension of $d_{model}$ through embedding, with a chosen value of 1024. The choice of $d_{model}$ The choice of $d_{model}$ is related to the richness of the model for detail learning, and a larger dmodel can contain more features.

Apart from the above adaption, other structure of the nets is basically constant with previous work. Transformer backbone employed in our study consisted of 6 layers in both the encoder and decoder and all attention mechanisms have 8 heads, which is a commonly used scale. During training, learning rate retain 0.0001 until 4,000 steps' warmup, then reduce linearly. Furthermore, the evaluation of the loss function was performed using the Mean Squared Error (MSE) criterion, and the activation function employed was Gelu, both of which are consistent with the commonly used advanced transformer applications. Our developed Transformer model has approximately 50.5 million(M) parameters.

The training process was executed on workstations equipped with Ascend AI processors. The network was developed using the PyTorch framework, and meticulous adjustments were made to the computing interface to accommodate Ascend's unique Compute Architecture for Neural Networks (CANN) heterogeneous computing architecture. This adaptation allowed the network to fully leverage the computational capabilities of Ascend processors during training. The training workload was efficiently distributed across 8 HUAWEI Ascend 910B cards, each equipped with 32GB of memory (amounting to a total of 256GB). This configuration enabled the utilization of a training batch size of 16384(with 2048 samples allocated to each card). Notably, this batch size represents the maximum capacity attainable for this hardware setup. The adoption of a larger batch size in our training was motivated by its potential benefits. A larger batch size facilitates the model's acquisition of general characteristics from the dataset, enabling it to identify the correct direction of optimization effectively. In contrast, smaller batch sizes might lead the model to focus excessively on the intricate features of a limited subset of data, potentially hindering convergence and impacting the overall training performance.

The experiment focuses on selecting typical multi-category expressway nodes and sections as learning resources. These resources encompass four sites, comprising five representative expressway scenarios, such as congestion, free flow, diversion, merge, weave, and their combinations. The trajectory data constructed for these sections is sourced from the UTE dataset (*23*), with the specific stations employed outlined in the **TABLE 1**. To ensure consistency and standardization, the data is first formatted and converted to a geodetic scale in meters. Subsequently, the data undergoes organization as detailed in the methodology. It is partitioned into areas with a length (L) of 300 meters and a width (W) of 20 meters. Furthermore, the upper limit of vehicles in each area is set at 10, aligning with the predefined network settings. Upon completion of these data preparation steps, the data utilized for training constitutes one epoch, spanning over 400,000 steps. To enhance model training and foster robustness, this data is iteratively used for pre-training across 50 epochs.

**TABLE 1 Training Dataset Consturcted Sources**

| Site | Traffic condition | Total Sampled |
|------|-------------------|---------------|
| KZM9 | Free flow and congestion in weaving section | |
| YTAvenue | Free flow in Merge/Diverge Segment | |
| KZM5 | Free flow in weaving section | 349,907 |
| KZM6 | Congestion in weaving section | |
| RML7 | Congestion and free flow in merging section | |

**RESULTS**
**Overall Performances**

In this section, we exam pre-training effect and employ a simple fine-tuning models to evaluate downstream useability. The fine-tuning task is trajectory compensation. In the specific implementation, [Mask] is used to cover some frames in the middle of the trajectory to form a gap. The masking method is the same as the noise addition method, and the goal of the model is to complete the prediction of these masking values. This fine-tuning trained for 15 epochs, about 7.2 million steps.

The test dataset consists of 10,000 samples, which share the same locations as the training dataset but are not included in the training set. The comprehensive evaluation of the predictions was performed using the Root Mean Square Error (RMSE) to compare the distance between predicted positions and ground truth values, as outlined in **Equation 2**. To ensure that the predicted vehicle distribution aligns with the real distribution, for instance, avoiding overlapping bounding boxes, the evaluation metrics include the Overlap Rate of predicted bounding boxes, representing the proportion of overlapping vehicles to the total number of vehicles, as defined in **Equation 3**. Moreover, the Intersection over Union (IoU) between the predicted values and ground truth is used to assess the consistency of bounding box positions and scales with the real values (**Equation 4**).

$$\text{RMSE} = \frac{1}{n}\sqrt{\sum_{i=1}^{n}\left[\left(X_{ipd} - X_{igt}\right)^2 + \left(Y_{ipd} - Y_{igt}\right)^2 + \left(W_{ipd} - W_{igt}\right)^2 + \left(H_{ipd} - H_{igt}\right)^2\right]} \quad (2)$$

$$\text{Overlap Rate} = \frac{Count_{overlap}}{Count_{all}} \tag{3}$$

$$IoU = \frac{area_{overlap}}{area_{union}} \tag{4}$$

where $X, Y, W, H$ indicates the bouding boxes positions and scales. $pd$ means prediction value and $gt$ is ground-truth.

In vehicle group prediction, the model needs to determine the number of effective vehicles within a group. To achieve this, the Delta Car Number (DCN) metric is introduced, which quantifies the absolute differences in the number of vehicles predicted by the model and the actual number of vehicles. Specifically, the pre-training task involves comparing the predictions for the next 10 frames with the ground truth values, while the evaluation for fine-tuning task entails comparing the generated values for the positions that need to be completed with the ground truth values. The detailed evaluation results are presented in **TABLE 2**.

**TABLE 2 Overall Evaluation**

| Model | Task | RMSE | Overlap Rate | IoU | DCN |
|-------|------|------|--------------|-----|-----|
| Pre-train | Prediction | 0.6059 | 0.8% | 0.56 | 90 |
| Fine-tune | Compensation | 0.3202 | 0.3% | 0.73 | 0 |

From the evaluation results, it can be observed that the pre-training task successfully learned the distribution pattern of vehicles, as there were almost no overlapping vehicles among the 10,000 data points in the test set. Additionally, the predicted vehicle positions closely matched the ground truth positions. The pre-training task achieved an RMSE of 0.6059 for vehicle positions, with an average Intersection over Union (IoU) of 0.56 for vehicle bounding boxes. The number of mis-predicted vehicle targets (excess or missing vehicles) was less than 1%. With the sustain of pre-trained model, fine-tune RMSE further decreased to 0.3202, while still maintaining the absence of overlap as in the pre-training task. The IoU between the completed bounding boxes and ground truth reached 0.73, indicating higher accuracy compared to the predictions. The DCN also reduced relatively. The excellent performance in the fine-tuning task may be attributed to the pre-training's learned features, and the task's ability to access trajectory information on both sides of the missing values, making the generation of missing points simpler compared to predictions with only one side's known information.

The overall predictions for various traffic scenarios are shown in **Figure 3**. Contains two-way road sections, diversion areas and merge areas, and traffic jams. We input the historical trajectory of the vehicle group into one-time prediction of the future operation of the group. It can be found that the model can accurately identify the running direction and future trajectory of individual vehicles in the group in different scenarios. **Figure 4(a)** provides visualizations of several vehicle groups' prediction results in pre-training predition, where the green boxes represent the ground truth and the red boxes indicate the predicted results. The vehicle groups include vehicles from two-directional roads, and the model accurately predicted the positions for both directional vehicles, confirming the reliability.

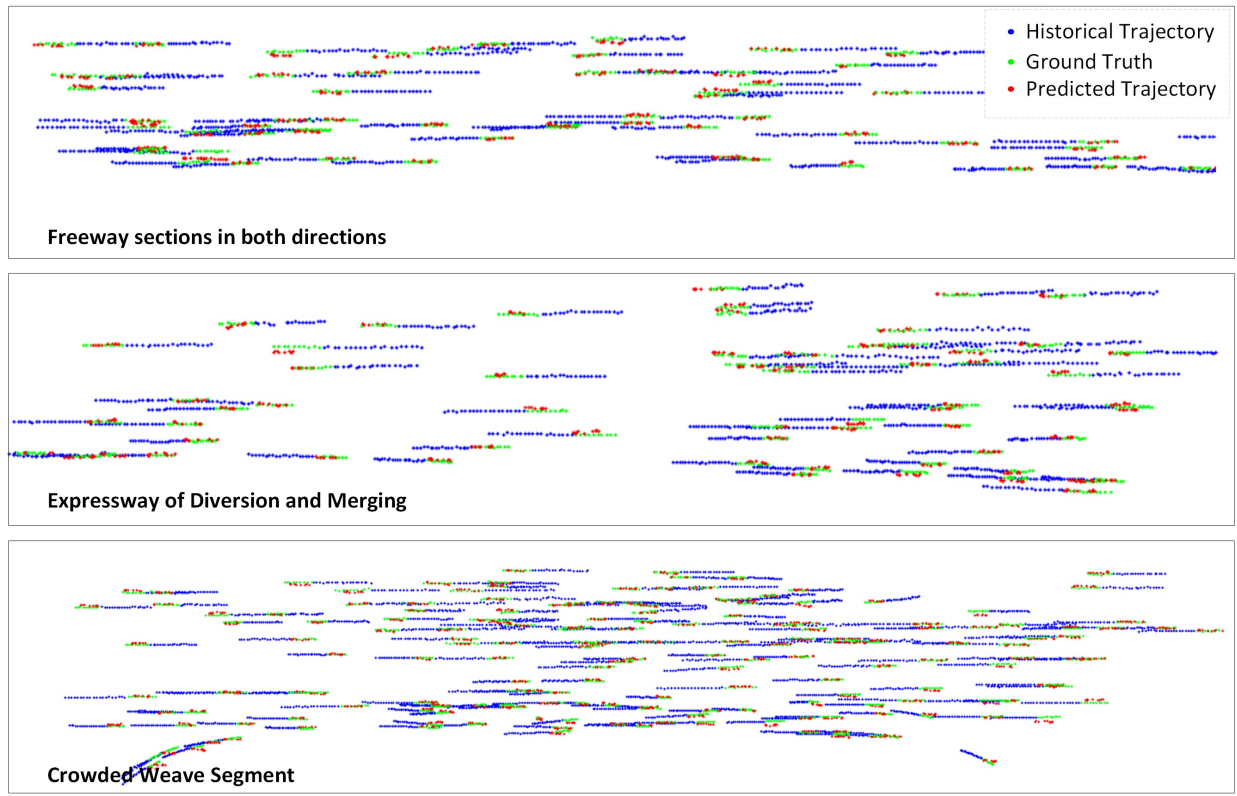

**Figure3** Overall view of trajectory prediction of vehicle groups in several scenarios.

In addition to position and scale accuracy, the reasonableness of vehicle velocities is also crucial. We measured the differences between the predicted and corresponding ground truth velocities for all trajectory predictions and plotted the frequency distribution of velocity differences, as shown in **Figure 4(b)**. The mean velocity difference is approximately 3 m/s, and 95% of velocity differences are below 7.5144 m/s. The model not only restored the positions of vehicle groups but also learned the operational speeds of the vehicle groups.

The trained model demonstrates excellent stability. During the training process, the prediction mode involves using 20 frames of historical trajectories with noise to predict the next 10 frames. To validate the stability of the model's predictions, we developed a continuous prediction mode. In this mode, the predicted trajectories are concatenated to the end of the current input and used as input for the next prediction. In other words, the input for the third iteration is the concatenation of the output values from the first and second iterations. In this continuous mode, we feed the concatenated predicted values as input for a loop of 20 times to examine whether errors amplify. We set the length of the predicted sequence to be 10 times the initial input length, i.e., Loop=20. Using 20 frames of input, the model predicts the trajectories of the vehicle groups for the next 200 frames.

The predicted trajectories of the vehicle groups are shown in **Figure 5**, where a 3D plot visualizes the trajectories of the vehicle groups along the lane and the vertical lane over time. Six vehicle groups were selected for illustration, and over the next 20 Loops, the trajectory segments seamlessly connect, with minimal velocity fluctuations. Even in scenarios involving lane changes, no trajectory overlaps were observed in the future. The predictions not only exhibit logical consistency but also reveal diverse driving behaviors among the vehicle groups. For instance, the propagation of traffic disturbances, deceleration for lane changes, and leading vehicles with free acceleration can be observed in the predicted trajectories. For the convenience of observation, two samples of driving behavior trajectories of deceleration and yielding are drawn in **Figure 6**. The dotted line represents the trajectory of other lanes, and the solid line represents the trajectory of the current lane. It can be seen from the trajectory space-time diagram that the simulated lane change process is very close to reality, and the following vehicle has a deceleration response to the inserted vehicle. Hence, the pretraining model has indeed learned the driving patterns of vehicles, demonstrating a variety of driving behaviors while ensuring stability and reasonableness.

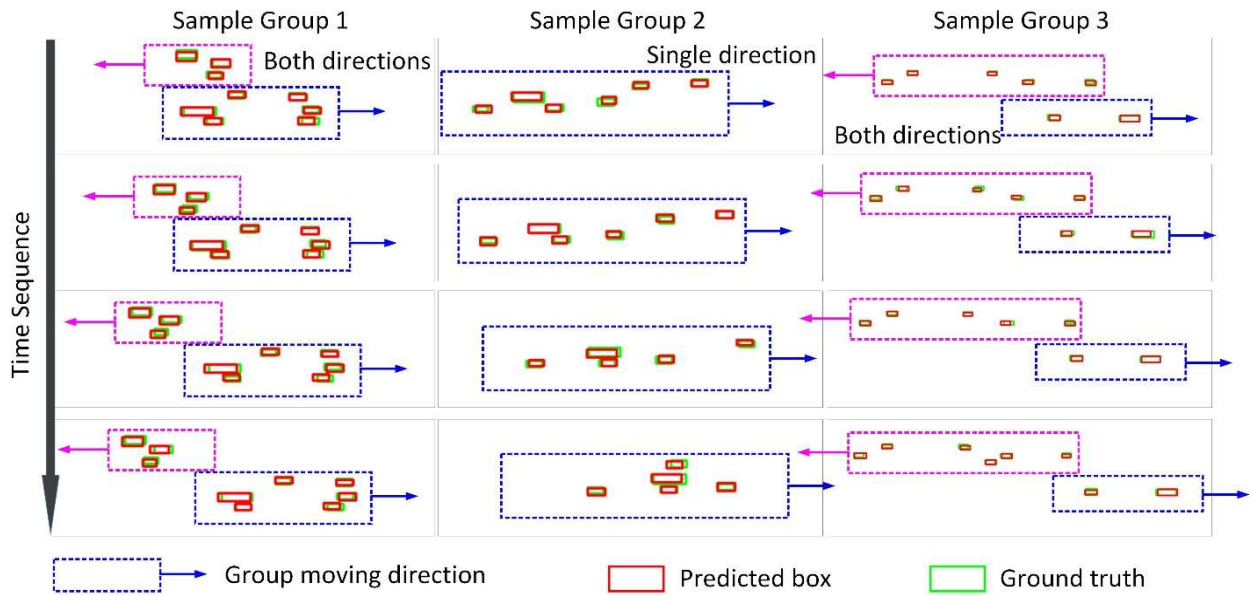

(b)

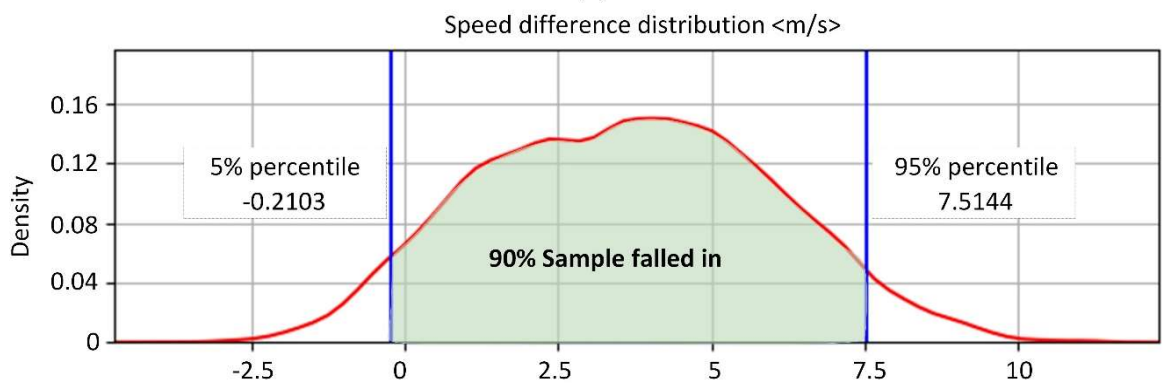

(c)

**Figure 4 Visualizations of vehicle groups' prediction and speed stability**

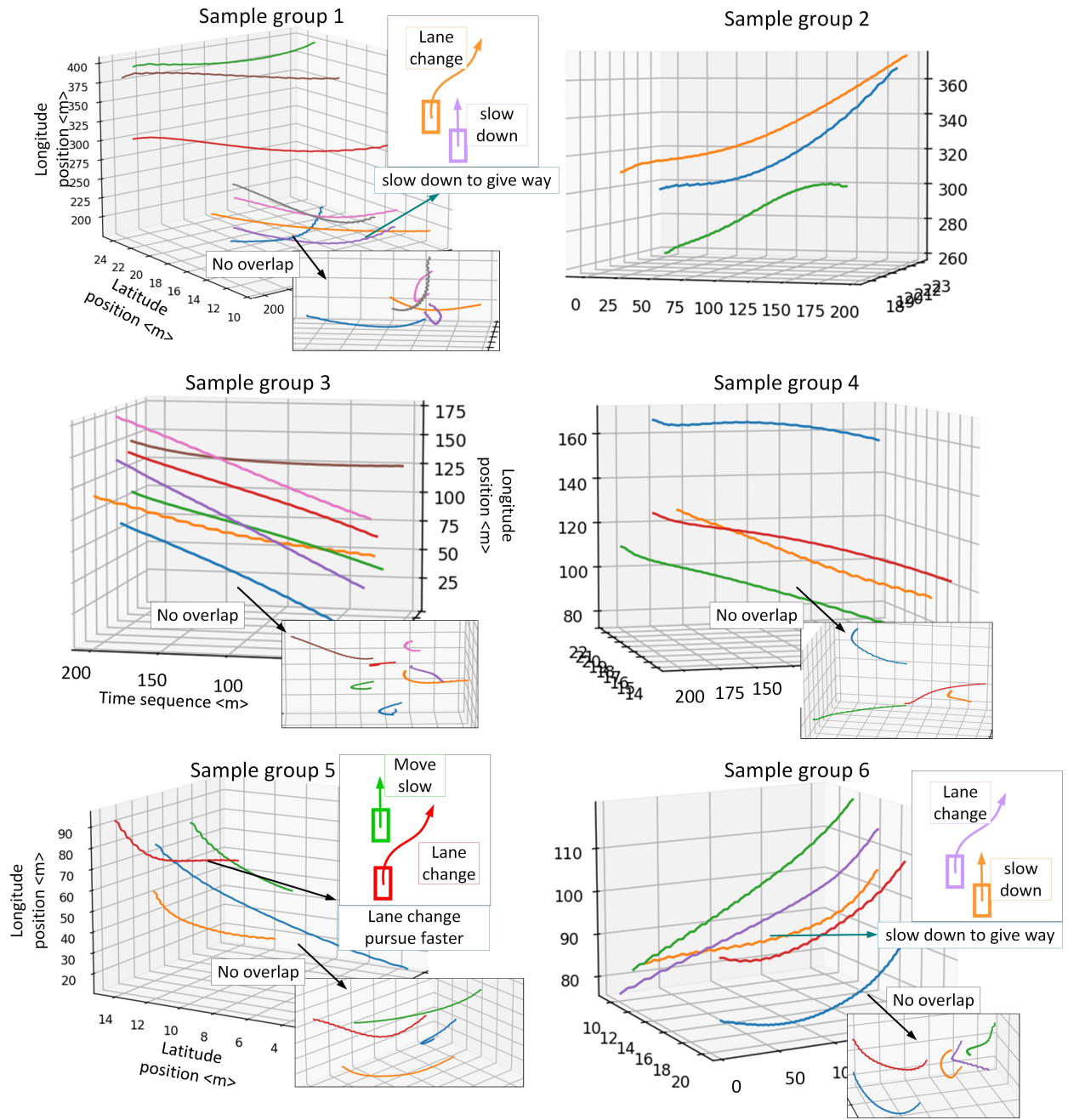

**Figure 5 Simulation of groups of vehicles for ten times the duration of the input**

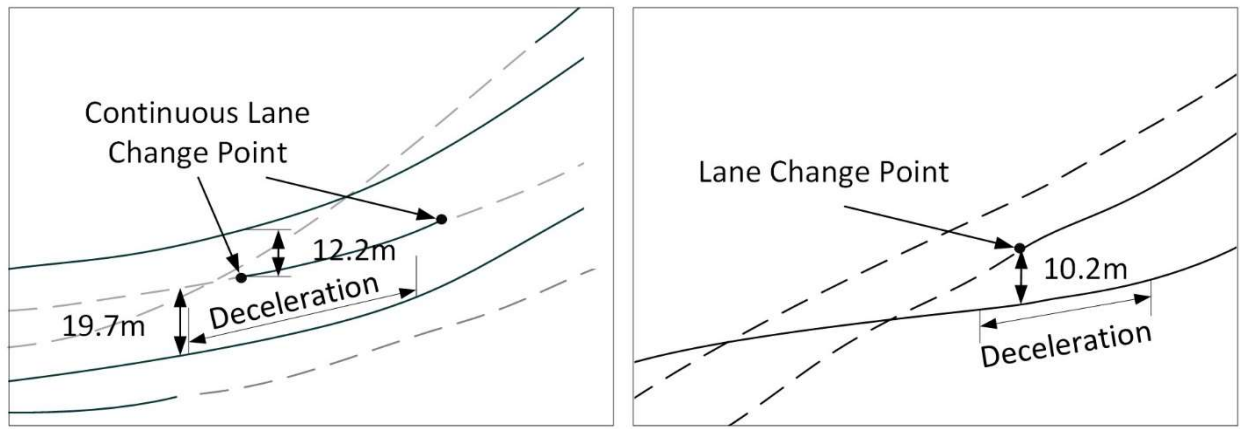

**Figure 6 Detailed display of driving behavior in continuous simulation**

## DISCUSSION AND CONCLUSIONS

This study presents a Transformer-based pre-training task specifically tailored to traffic scenarios, focusing on vehicle trajectory data as its core. The research dissects two sub-goals of traffic tasks and devises corresponding pre-training task patterns. Through these carefully designed tasks, the pre-trained model captures the intricate interactions inherent in vehicle driving processes, encompassing various driving patterns and heterogeneous behaviors. Verification is evaluated by RMSE and generating an overlap of the vehicle's bounding box. For the pre-trained model, the generated results are 99% avoiding vehicle overlap. This indicates that the model learns the distribution and interaction properties of vehicle boundaries. Compared with the true value, the pre-trained prediction results show an RMSE of 0.6059m, and 95% of which with speed deivation within 7.5144m/s. When adaption in further finetune task, better proformance certify the valued pre-trained model contribution.

It is essential to mention that training large-scale models poses a challenging and critical engineering problem, necessitating careful consideration of distributed design and implementation for multi-machine, multi-GPU parallelism. It is crucial to ensure a balanced utilization of resources on each GPU. Although our work benefits from numerical prediction, reducing the $d_{model}$ during training, it is objectively constrained by the available resources. The $8 \times 32GB$ memory on the GPUs is still insufficient to train a truly large-scale model. Consequently, this method did not incorporate longer input sequences. However, increasing the sequence length, for example, from 20 frames to hundreds of frames or more, could potentially lead to a more refined model and contribute to the stability of continuous predictions. However, such an extension would demand larger $d_{model}$, sequence length, and corresponding multiplied Datasets and memory resources, which currently exceed the available research resources.

We believe that this is a work with great potential, and it still needs to be further expanded. For future work, a richer pre-training model, such as extend sequence length, will be extended on the basis of this research to learn more diverse vehicle features and connect to more complex downstream tasks. In addition to the development of the pre-trained model, the testing of downstream tasks will be further refined. We plan to extend larger sequence length to fit more complex downstream tasks, such as traffic control, reproduction of regional complex traffic patterns. It also necessary to anticipate addressing biases in long-term trajectory predictions, evaluating the robustness of overall vehicle group simulations, and assessing whether long-distance simulations align with classical car-following models. These evaluations hold great importance at the application level and will be pursued in further research efforts.

## ACKNOWLEDGMENTS

We would like to express our gratitude for the financial support provided by the National Natural Science Foundation of China under Grant 52272331. Additionally, we extend our appreciation to HUAWEI Ascend for providing valuable device support, which enabled us to expand the scale of the experiment, prompting the effect of the model. Thanks to Yufan Wang and Ding Ma for their technology assistance. We are also indebted to Shunchao Wang for their exceptional management throughout the course of this study. Thanks to ChatGPT for helping us correct grammatical errors, which greatly improved our writing efficiency. Their support and expertise have been invaluable to the overall success of this research endeavor. Thanks for every gorgeous sunset and clear starry sky.

**AUTHOR CONTRIBUTIONS**
The authors confirm contribution to the paper as follows: study conception and design: R. Feng, Y. Ding and Z. Li; data collection: R. Feng; analysis and interpretation of results: R. Feng and B. Liu; draft manuscript preparation: R. Feng., and Z. Li. All authors reviewed the results and approved the final version of the manuscript.

**REFERENCES**

1. Coifman B. Empirical flow-density and speed-spacing relationships: Evidence of vehicle length dependency[J]. Transportation Research Part B: Methodological, 2015, 78: 54-65.

2. Siqueira A F, Peixoto C J T, Wu C, et al. Effect of stochastic transition in the fundamental diagram of traffic flow[J]. Transportation Research Part B: Methodological, 2016, 87: 1-13.

3. Zhou M, Qu X, Li X. A recurrent neural network based microscopic car following model to predict traffic oscillation[J]. Transportation research part C: emerging technologies, 2017, 84: 245-264.

4. Rhoades C, Wang X, Ouyang Y. Calibration of nonlinear car-following laws for traffic oscillation prediction[J]. Transportation research part C: emerging technologies, 2016, 69: 328-342.

5. Chiabaut N, Leclercq L, Buisson C. From heterogeneous drivers to macroscopic patterns in congestion[J]. Transportation Research Part B: Methodological, 2010, 44(2): 299-308.

6. Newell G F. A simplified car-following theory: a lower order model[J]. Transportation Research Part B: Methodological, 2002, 36(3): 195-205.

7. Chen D, Laval J, Zheng Z, et al. A behavioral car-following model that captures traffic oscillations[J]. Transportation research part B: methodological, 2012, 46(6): 744-761.

8. Chen D, Laval J A, Ahn S, et al. Microscopic traffic hysteresis in traffic oscillations: A behavioral perspective[J]. Transportation Research Part B: Methodological, 2012, 46(10): 1440-1453.

9. Sharma A, Zheng Z, Bhaskar A. Is more always better? The impact of vehicular trajectory completeness on car-following model calibration and validation[J]. Transportation research part B: methodological, 2019, 120: 49-75.

10. Liu Z, He J, Zhang C, et al. Vehicle trajectory extraction at the exit areas of urban freeways based on a novel composite algorithms framework[J]. Journal of Intelligent Transportation Systems, 2023, 27(3): 295-313.

11. Y. Liu, X. Wang, L. Li, S. Cheng and Z. Chen, "A Novel Lane Change Decision-Making Model of Autonomous Vehicle Based on Support Vector Machine," in IEEE Access, vol. 7, pp. 26543-26550, 2019, doi: 10.1109/ACCESS.2019.2900416.

12. Hu Y, Li Y, Huang H, et al. A high-resolution trajectory data driven method for real-time evaluation of traffic safety[J]. Accident Analysis & Prevention, 2022, 165: 106503.

13. Mo Z, Shi R, Di X. A physics-informed deep learning paradigm for car-following models[J]. Transportation research part C: emerging technologies, 2021, 130: 103240.

14. Ma L, Qu S. A sequence to sequence learning based car-following model for multi-step predictions considering reaction delay[J]. Transportation research part C: emerging technologies, 2020, 120: 102785.

15. Vaswani A, Shazeer N, Parmar N, et al. Attention is all you need[J]. Advances in neural information processing systems, 2017, 30.

16. Brown T, Mann B, Ryder N, et al. Language models are few-shot learners[J]. Advances in neural information processing systems, 2020, 33: 1877-1901.

17. Devlin J, Chang M W, Lee K, et al. Bert: Pre-training of deep bidirectional transformers for language understanding[J]. arXiv preprint arXiv:1810.04805, 2018.

18. Lewis M, Liu Y, Goyal N, et al. Bart: Denoising sequence-to-sequence pre-training for natural language generation, translation, and comprehension[J]. arXiv preprint arXiv:1910.13461, 2019.

19. Raffel C, Shazeer N, Roberts A, et al. Exploring the limits of transfer learning with a unified text-to-text transformer[J]. The Journal of Machine Learning Research, 2020, 21(1): 5485-5551.

20. Radford A, Narasimhan K, Salimans T, et al. Improving language understanding by generative pre-training[J]. 2018.

21. Bao H, Dong L, Piao S, et al. Beit: Bert pre-training of image transformers[J]. arXiv preprint arXiv:2106.08254, 2021.

22. He K, Chen X, Xie S, et al. Masked autoencoders are scalable vision learners[C]//Proceedings of the IEEE/CVF conference on computer vision and pattern recognition. 2022: 16000-16009.

23. Feng R, Fan C, Li Z, et al. Vehicle Trajectory Construction Framework from Aerial Videos based on Convolution Neural Network Detection[C]. Transportation Research Board Annual Meeting, 2020.